\documentclass[sigconf]{acmart}

\usepackage[ruled,linesnumbered]{algorithm2e}

\AtBeginDocument{%
  }

\setcopyright{none}

\renewcommand\footnotetextcopyrightpermission[1]{}
\begin{document}

\fancyhf{}
\renewcommand{\headrulewidth}{0pt}
\renewcommand{\footrulewidth}{0pt}

\title{Fraud Detection Through Large-Scale Graph Clustering with Heterogeneous Link Transformation}

\author{Chi Liu}

\begin{abstract}
Collaborative fraud, where multiple fraudulent accounts coordinate to exploit online payment systems, poses significant challenges due to the formation of complex network structures. Traditional detection methods that rely solely on high-confidence identity links suffer from limited coverage, while approaches using all available linkages often result in fragmented graphs with reduced clustering effectiveness. In this paper, we propose a novel graph-based fraud detection framework that addresses the challenge of large-scale heterogeneous graph clustering through a principled link transformation approach. Our method distinguishes between \emph{hard links} (high-confidence identity relationships such as phone numbers, credit cards, and national IDs) and \emph{soft links} (behavioral associations including device fingerprints, cookies, and IP addresses). We introduce a graph transformation technique that first identifies connected components via hard links, merges them into super-nodes, and then reconstructs a weighted soft-link graph amenable to efficient embedding and clustering. The transformed graph is processed using LINE (Large-scale Information Network Embedding) for representation learning, followed by HDBSCAN (Hierarchical Density-Based Spatial Clustering of Applications with Noise) for density-based cluster discovery. Experiments on a real-world payment platform dataset demonstrate that our approach achieves significant graph size reduction (from 25 million to 7.7 million nodes), doubles the detection coverage compared to hard-link-only baselines, and maintains high precision across identified fraud clusters. Our framework provides a scalable and practical solution for industrial-scale fraud detection systems.
\end{abstract}

\ccsdesc[500]{Security and privacy~Intrusion/anomaly detection and malware mitigation}
\ccsdesc[500]{Information systems~Clustering}
\ccsdesc[300]{Computing methodologies~Machine learning approaches}

\keywords{fraud detection, graph clustering, graph embedding, heterogeneous networks, payment security}

\maketitle

\section{Introduction}

\begin{figure}[t]
  \centering
  \includegraphics[width=\columnwidth]{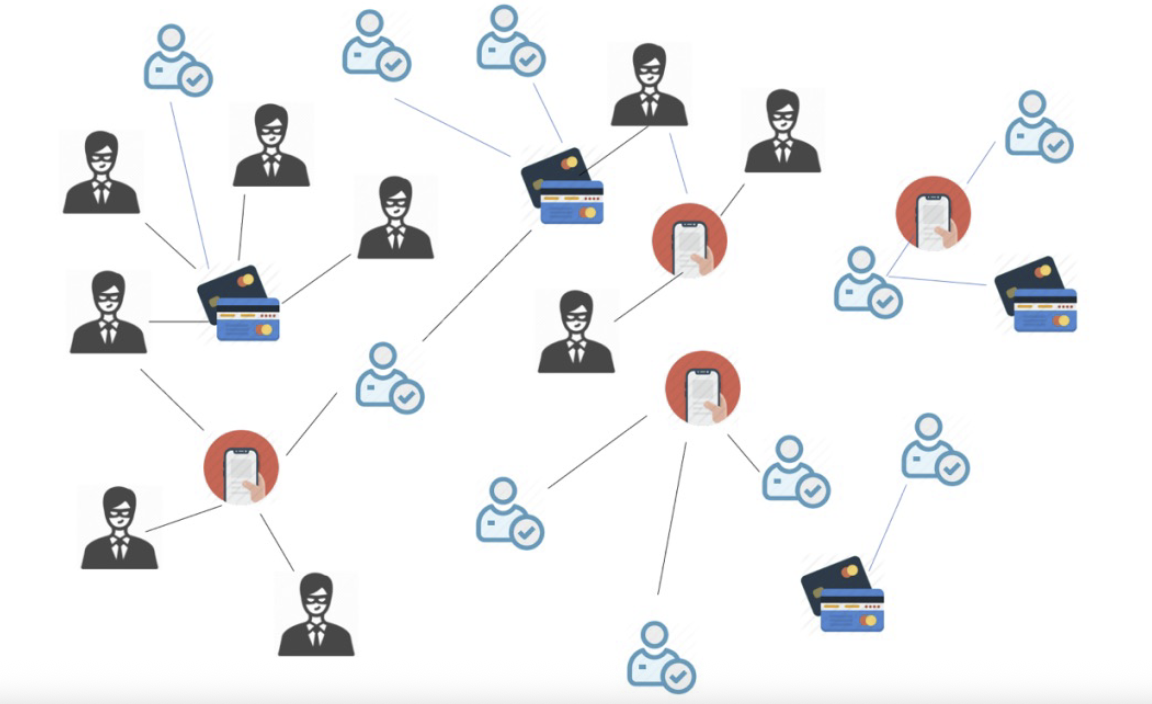}
  \caption{Illustration of a fraud network. Fraudulent actors (dark figures) and legitimate users (blue figures with checkmarks) are connected through shared devices (phones with red backgrounds) and payment instruments (credit cards). These connections form implicit networks that can be detected through graph-based analysis.}
  \Description{Network diagram showing dark figures representing fraudsters and blue figures with checkmarks representing legitimate users, connected by phones in red circles and credit cards.}
  \label{fig:fraud-network}
\end{figure}

Online payment platforms process billions of transactions daily, making them attractive targets for fraudulent activities. Among the most challenging threats is \emph{collaborative fraud}, where multiple seemingly independent accounts coordinate to exploit platform vulnerabilities, share stolen credentials, or conduct organized financial crimes~\cite{akoglu2015graph}. These fraudulent actors often form implicit networks through shared identifiers, devices, and behavioral patterns, creating opportunities for graph-based detection approaches.

Traditional fraud detection systems typically rely on rule-based methods or supervised machine learning models that analyze individual account features~\cite{bolton2002statistical}. While effective for isolated fraud cases, these approaches struggle to detect collaborative fraud patterns where individual accounts may appear legitimate but collectively exhibit suspicious behavior. Graph-based methods offer a promising alternative by modeling relationships between accounts and identifying anomalous network structures~\cite{pourhabibi2020fraud}. Graph clustering, which aims to partition nodes into homogeneous groups based on connectivity patterns, has emerged as a powerful technique for uncovering hidden community structures~\cite{fortunato2010community}, with applications spanning social network analysis, bioinformatics, and fraud detection~\cite{watteau2024graph}.

However, applying graph-based fraud detection at industrial scale presents significant challenges. First, the \textbf{scale} of the problem is enormous---modern payment platforms maintain graphs with tens of millions of nodes and hundreds of millions of edges. Second, the graphs are inherently \textbf{heterogeneous}, containing diverse types of linkages with varying reliability levels. For instance, two accounts sharing a verified phone number likely belong to the same user, while accounts sharing an IP address may simply be using the same public network. Third, these graphs are often \textbf{sparse} and \textbf{fragmented}, with meaningful clusters separated by weak or noisy connections. Finally, the \textbf{unsupervised} nature of the problem---where labeled fraud clusters are scarce---necessitates methods that can discover suspicious patterns without extensive ground truth. While recent graph neural networks (GNNs) have shown promise in fraud detection~\cite{liu2018heterogeneous}, they typically require labeled training data, struggle with scalability on billion-edge graphs due to neighborhood aggregation overhead, and demand careful hyperparameter tuning. These constraints motivate our focus on efficient unsupervised methods that can scale to industrial datasets.

Existing graph-based approaches face a fundamental trade-off. Methods that rely solely on high-confidence linkages (e.g., shared payment credentials) achieve high precision but suffer from limited coverage, as many fraud rings deliberately avoid reusing such identifiers. Conversely, approaches that incorporate all available linkages often produce overly connected graphs where meaningful cluster structures are obscured by noise, or excessively fragmented graphs where important relationships are lost.

In this paper, we propose a novel framework that addresses these challenges through a principled \emph{heterogeneous link transformation} approach. Our key insight is that different types of linkages serve fundamentally different purposes in fraud network analysis. We categorize linkages into two classes:
\begin{itemize}
    \item \textbf{Hard links}: High-confidence identity relationships (phone numbers, credit cards, email addresses, national IDs, bank accounts) that strongly suggest account ownership by the same entity.
    \item \textbf{Soft links}: Behavioral associations (device fingerprints, cookies, IP addresses) that indicate potential relationships but may also arise from coincidental sharing.
\end{itemize}

\begin{figure*}[t]
  \centering
  \includegraphics[width=0.95\textwidth]{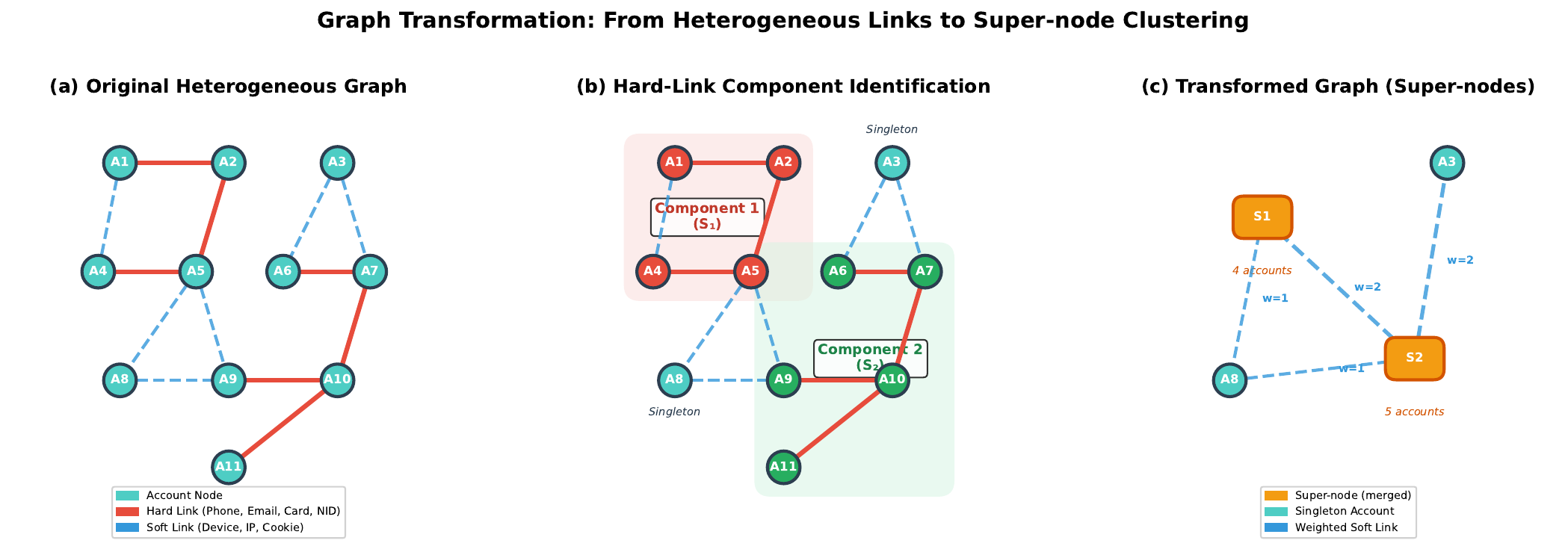}
  \caption{Overview of the proposed graph transformation framework. (a) Original heterogeneous graph with account nodes (teal circles) connected by hard links (red solid lines: phone, email, credit card, national ID) and soft links (blue dashed lines: device fingerprint, IP address, cookie). (b) Hard-link connected components are identified: Component 1 ($S_1$, red background) containing accounts A1, A2, A4, A5, and Component 2 ($S_2$, green background) containing accounts A6, A7, A9, A10, A11. Singletons A3 and A8 have no hard links. (c) Transformed graph where components are merged into super-nodes (orange rounded rectangles), connected only by weighted soft links. Edge weights reflect the number of aggregated soft links between super-nodes.}
  \Description{A three-part diagram showing graph transformation: (a) original graph with 11 account nodes in teal connected by red hard links and blue dashed soft links, (b) intermediate step showing two hard-link components identified with colored backgrounds, (c) final transformed graph with 2 orange super-nodes, 2 singleton accounts, and weighted soft-link edges.}
  \label{fig:overview}
\end{figure*}

Our framework first identifies connected components through hard links and merges them into \emph{super-nodes}, effectively consolidating accounts that almost certainly belong to the same entity. We then reconstruct a weighted graph using soft links between super-nodes, where edge weights reflect the strength of association. This transformed graph is substantially smaller and exhibits clearer cluster structures, making it amenable to efficient embedding and clustering algorithms. Figure~\ref{fig:overview} illustrates this transformation process.

We apply LINE (Large-scale Information Network Embedding)~\cite{tang2015line} to learn low-dimensional representations of super-nodes that preserve both first-order and second-order proximity. The resulting embeddings are then clustered using HDBSCAN~\cite{campello2013density}, a density-based algorithm that automatically identifies clusters of varying densities without requiring a predefined number of clusters---a crucial property for fraud detection where the number and size of fraud rings are unknown.

We evaluate our framework on a large-scale real-world dataset from a major payment platform. Our experiments demonstrate that the graph transformation achieves significant reduction in graph size (from 25 million to 7.7 million nodes), enabling efficient processing. Compared to hard-link-only baselines, our approach doubles the detection coverage while maintaining high precision. The framework has been deployed in production and successfully identifies previously undetected fraud rings.

\vspace{0.5em}
\noindent\textbf{Contributions.} Our main contributions are:
\begin{itemize}
    \item We formalize the distinction between hard links and soft links in fraud detection graphs and propose a principled graph transformation technique that leverages this distinction to reduce graph complexity while preserving meaningful structures.
    \item We present an end-to-end framework combining graph transformation, network embedding, and density-based clustering for large-scale unsupervised fraud detection.
    \item We provide extensive experimental evaluation on real-world data, demonstrating significant improvements in both scalability and detection coverage over existing approaches.
    \item We share practical insights from deploying the system in a production environment serving hundreds of millions of users.
\end{itemize}

\section{Related Work}

Our work relates to three main areas: graph-based fraud detection, network embedding, and graph clustering. We discuss each in turn.

\subsection{Graph-based Fraud Detection}

Fraud detection has evolved from rule-based systems~\cite{bolton2002statistical} to sophisticated machine learning approaches. Graph-based methods have gained prominence due to their ability to capture relational patterns among entities~\cite{akoglu2015graph}. Early work focused on detecting anomalous subgraphs~\cite{noble2003graph} and identifying suspicious dense subgraphs in bipartite networks~\cite{hooi2016fraudar}. More recent approaches leverage graph neural networks for fraud detection~\cite{liu2018heterogeneous}, learning node representations that incorporate both structural and attribute information.

GNN-based fraud detection methods typically operate in a semi-supervised or supervised setting, requiring labeled fraud examples for training. While effective when labels are abundant, they face challenges in real-world fraud detection: (1) labeled fraud data is scarce and biased toward already-detected patterns, (2) fraudsters constantly evolve tactics, rendering historical labels less informative, and (3) the labeling process itself is expensive and slow. These limitations motivate \textit{unsupervised} approaches that can discover novel fraud patterns without training labels.

A key challenge in fraud detection is the heterogeneous nature of entity relationships. Fraudsters deliberately create complex networks mixing high-confidence identity links with behavioral associations to evade detection~\cite{pourhabibi2020fraud}. Our work addresses this by explicitly distinguishing between hard links (identity-based) and soft links (behavior-based), enabling more effective preprocessing before applying unsupervised embedding and clustering techniques.

\subsection{Network Embedding}

Network embedding methods learn low-dimensional representations of nodes that preserve network structure. Classical approaches such as Laplacian Eigenmaps~\cite{belkin2001laplacian} and IsoMap~\cite{tenenbaum2000global} compute eigenvectors of affinity matrices but scale poorly to large networks.

DeepWalk~\cite{perozzi2014deepwalk} pioneered scalable network embedding by using truncated random walks and skip-gram models. LINE~\cite{tang2015line} introduced explicit objectives for preserving first-order proximity (direct connections) and second-order proximity (shared neighborhoods). Node2Vec~\cite{grover2016node2vec} extended random walk strategies with biased sampling to balance breadth-first and depth-first exploration. Graph neural networks such as GraphSAGE~\cite{hamilton2017inductive} and GCN~\cite{kipf2016semi} learn embeddings through neighborhood aggregation.

Our framework builds upon LINE for its efficiency on billion-edge graphs and its principled handling of weighted edges. However, we introduce a crucial preprocessing step---graph transformation via hard-link merging---that significantly reduces graph complexity while preserving meaningful structures for downstream embedding.

\subsection{Graph Clustering}

Graph clustering aims to partition nodes into groups with dense internal connections~\cite{schaeffer2007graph}. A comprehensive survey of graph clustering methods can be found in~\cite{watteau2024graph}. Traditional methods include spectral clustering~\cite{von2007tutorial}, which leverages eigenvalues and eigenvectors of the graph Laplacian matrix to identify cluster structures, and modularity optimization approaches such as the Louvain~\cite{blondel2008fast} and Leiden~\cite{traag2019leiden} algorithms. Stochastic Block Models (SBM)~\cite{holland1983stochastic} provide a probabilistic framework for community detection, where nodes are partitioned into blocks with probabilistic connection patterns. The Markov Clustering Algorithm~\cite{van2008graph} uses random walks and inflation operations to identify densely connected regions.

Recent advances in deep learning have led to the emergence of deep graph clustering methods~\cite{liu2022deep}. Graph Autoencoders (GAE)~\cite{kipf2016variational} encode graph structures into latent representations suitable for clustering. Adversarially Regularized Graph Autoencoders (ARGA)~\cite{pan2018adversarially} improve upon GAE by incorporating adversarial training for better latent space regularization. Contrastive learning approaches such as MVGRL~\cite{hassani2020contrastive} maximize mutual information between different views of the graph to learn robust representations.

Density-based clustering algorithms such as DBSCAN~\cite{ester1996density} and its hierarchical extension HDBSCAN~\cite{campello2013density} are particularly suitable for fraud detection scenarios where the number of fraud rings is unknown and clusters may have varying densities. HDBSCAN automatically determines the number of clusters and handles noise points, making it robust for real-world fraud networks where many accounts may not belong to any fraud ring.

Our contribution lies not in proposing new embedding or clustering algorithms, but in developing a principled graph transformation technique that bridges heterogeneous link types with these existing methods. By merging hard-link components into super-nodes before embedding, we achieve both computational efficiency and improved cluster quality. This preprocessing step is complementary to both traditional methods (spectral clustering, modularity optimization) and deep learning approaches (GAE, GNN-based methods), enabling their application to graphs that would otherwise be too large or too noisy to process effectively

\section{Problem Formulation}
\label{sec:problem}

We formally define the problem of fraud detection through heterogeneous graph clustering. We begin by defining the heterogeneous account graph and the two types of linkages, then formulate the graph transformation and fraud cluster discovery problems. A summary of notation is provided in Table~\ref{tab:notation}.

\subsection{Heterogeneous Account Graph}

\begin{definition}
\label{def:account_graph}
\textbf{(Heterogeneous Account Graph)}
\textsl{A heterogeneous account graph is defined as $G = (V, E_H, E_S)$, where $V$ is the set of account nodes, $E_H$ is the set of \textbf{hard-link} edges, and $E_S$ is the set of \textbf{soft-link} edges. Each edge $e \in E_H \cup E_S$ connects two accounts $(u, v)$ that share a common identifier or behavioral pattern.}
\end{definition}

\begin{figure}[t]
  \centering
  \includegraphics[width=\columnwidth]{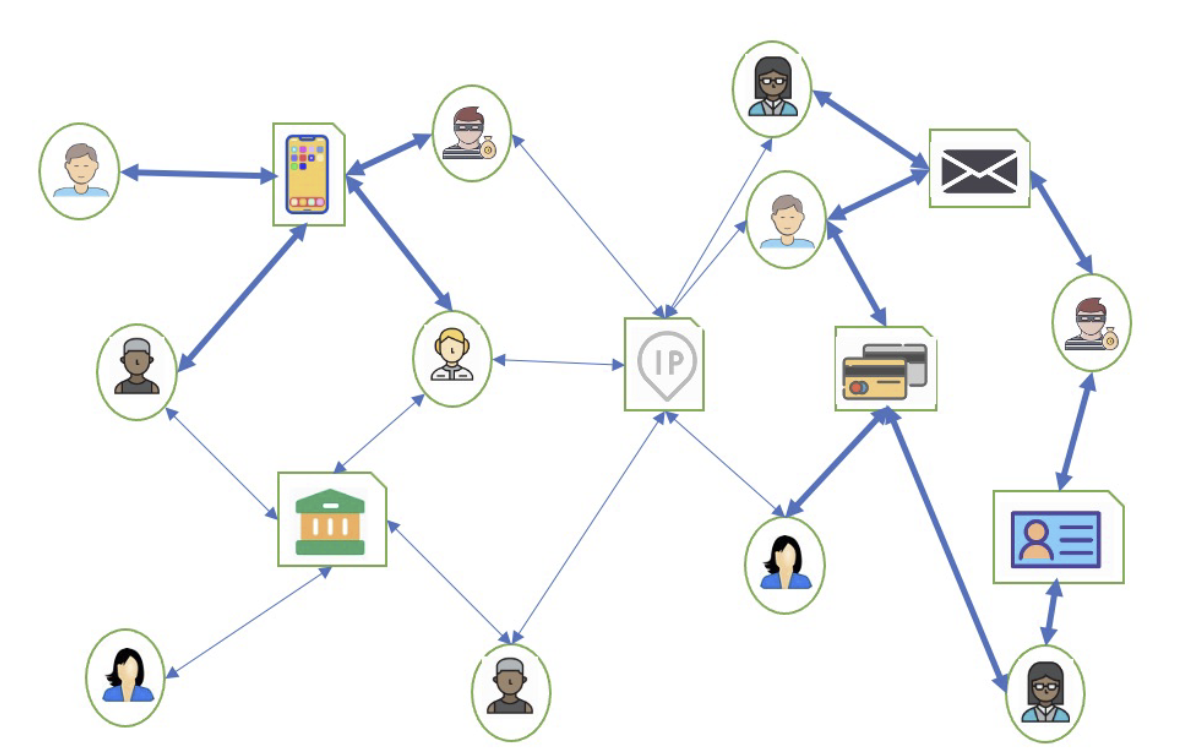}
  \caption{Example of a heterogeneous account graph. User accounts (circles with green borders) are connected to various entities including phones, emails, IP addresses, credit cards, bank accounts, and ID documents (shown as boxes with green borders). Blue arrows represent strong connections between accounts and entities, while grey lines indicate weaker behavioral associations.}
  \Description{Network graph showing user account nodes connected to entity boxes representing phones, emails, IPs, cards, banks, and IDs.}
  \label{fig:hetero-graph}
\end{figure}

In practice, account graphs in payment platforms can contain tens of millions of nodes and hundreds of millions of edges. The heterogeneous nature of links---with varying reliability and semantic meaning---distinguishes this setting from standard homogeneous graph problems.

\subsection{Hard Links and Soft Links}

\begin{figure}[t]
  \centering
  \includegraphics[width=\columnwidth]{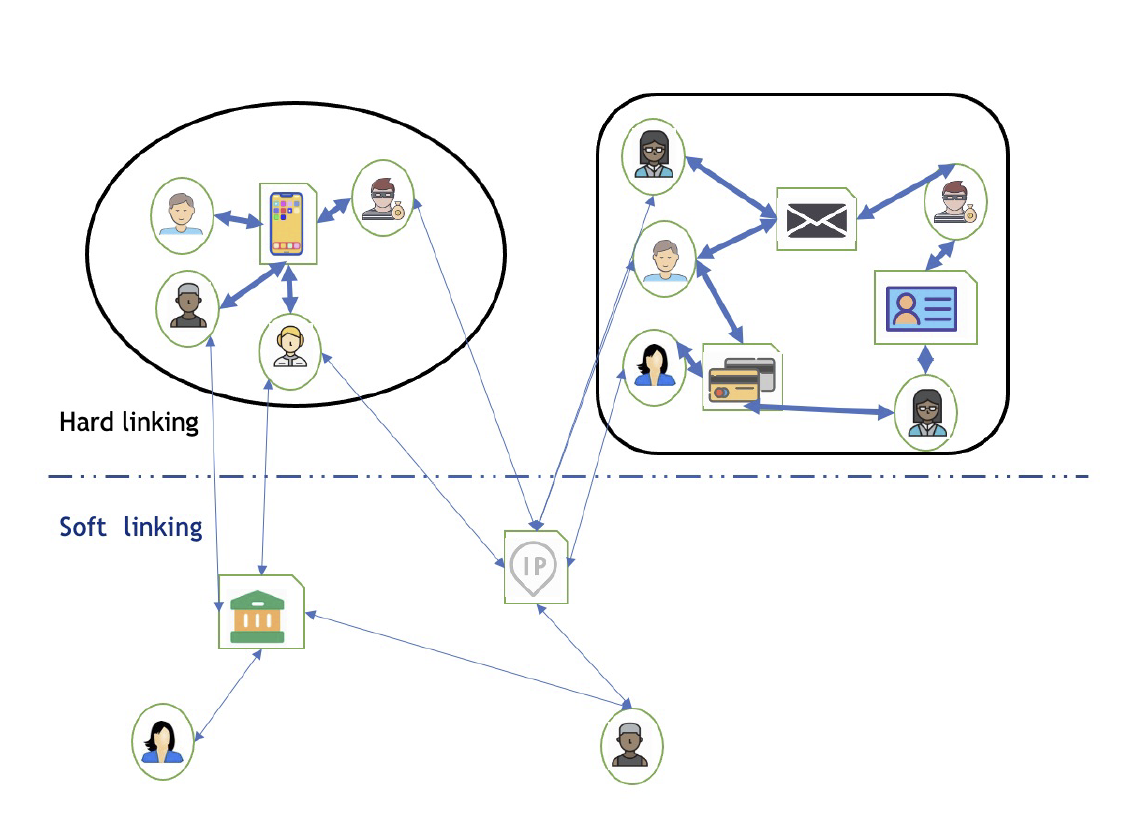}
  \caption{Illustration of hard links vs. soft links in the graph transformation process. \textbf{Upper region (Hard-Link Component 1):} Accounts A, B, and C are connected by hard links (sharing verified credentials: phone, email, card). These form a tightly connected component that will merge into a single super-node. \textbf{Lower region (Hard-Link Component 2):} Accounts D and E share hard links (ID, bank account), forming another super-node. \textbf{Cross-component soft links (dashed lines):} Behavioral associations (shared device fingerprint between A and D; shared IP address between C and E) connect accounts from different hard-link components. After transformation, these soft links become weighted edges between the two resulting super-nodes. \textbf{Within-component soft links:} If any soft link exists between accounts within the same hard-link component (not shown), it would be discarded as redundant since those accounts already merge into one super-node.}
  \Description{Diagram showing two hard-link components in bounded regions connected by soft links across a dashed horizontal line.}
  \label{fig:hard-soft-concept}
\end{figure}

The key insight of our approach is the principled distinction between two fundamentally different types of linkages (illustrated in Figure~\ref{fig:hard-soft-concept}):

\begin{definition}
\label{def:hard_link}
\textbf{(Hard Link)}
\textsl{A \textbf{hard link} $(u, v) \in E_H$ represents a high-confidence identity relationship between accounts $u$ and $v$. Hard links are established when accounts share verified identity credentials including: phone numbers, email addresses, credit card numbers, national identification numbers, or bank account numbers. The presence of a hard link strongly suggests that the two accounts are controlled by the same entity.}
\end{definition}

\begin{definition}
\label{def:soft_link}
\textbf{(Soft Link)}
\textsl{A \textbf{soft link} $(u, v) \in E_S$ represents a behavioral association between accounts $u$ and $v$. Soft links are established when accounts share: device fingerprints, browser cookies, IP addresses, or other behavioral identifiers. Each soft link $e \in E_S$ is associated with a weight $w_e > 0$ indicating the strength of association, which may be computed based on frequency of co-occurrence or other similarity measures.}
\end{definition}

The distinction between hard and soft links reflects their different roles in fraud network analysis. Hard links provide high-precision identity consolidation but may have limited coverage, as sophisticated fraudsters deliberately avoid sharing verified credentials across accounts. Soft links provide broader coverage of potential relationships but may include false positives due to coincidental sharing (e.g., multiple users on the same public WiFi network).

\subsection{Super-Node Transformation}

To leverage the complementary strengths of hard and soft links, we define a graph transformation that consolidates hard-link connected components into super-nodes:

\begin{definition}
\label{def:super_node}
\textbf{(Super-Node)}
\textsl{Given a heterogeneous account graph $G = (V, E_H, E_S)$, a \textbf{super-node} $S_i$ is a maximal connected component in the subgraph induced by hard links $(V, E_H)$. That is, $S_i \subseteq V$ such that for any two accounts $u, v \in S_i$, there exists a path of hard links connecting them, and no account in $S_i$ has a hard link to any account outside $S_i$.}
\end{definition}

\begin{definition}
\label{def:transformed_graph}
\textbf{(Transformed Graph)}
\textsl{The \textbf{transformed graph} $G' = (V', E')$ is constructed from $G = (V, E_H, E_S)$ as follows:
\begin{itemize}
    \item $V' = \{S_1, S_2, \ldots, S_k\}$ is the set of super-nodes, where each $S_i$ is a hard-link connected component.
    \item $E'$ contains a weighted edge $(S_i, S_j)$ for each pair of super-nodes connected by at least one soft link. The edge weight $w_{ij}$ aggregates the soft-link weights between accounts in the two super-nodes:
    \begin{equation}
    w_{ij} = \sum_{u \in S_i, v \in S_j, (u,v) \in E_S} w_{(u,v)}
    \end{equation}
\end{itemize}}
\end{definition}

This transformation achieves two objectives: (1) it reduces the graph size by consolidating accounts that almost certainly belong to the same entity, and (2) it produces a weighted graph where edge weights reflect the aggregate strength of behavioral associations between entity groups.

\subsection{Problem Statement}

With the above definitions, we can formally state our problem:

\begin{definition}
\label{def:problem}
\textbf{(Fraud Cluster Discovery)}
\textsl{Given a heterogeneous account graph $G = (V, E_H, E_S)$, the problem of \textbf{fraud cluster discovery} aims to:
\begin{enumerate}
    \item Construct the transformed graph $G' = (V', E')$ through super-node merging.
    \item Learn a low-dimensional embedding $\phi: V' \rightarrow \mathbb{R}^d$ that preserves the structural properties of $G'$, where $d \ll |V'|$.
    \item Identify clusters $\mathcal{C} = \{C_1, C_2, \ldots, C_m\}$ in the embedding space that correspond to potential fraud rings, where each cluster $C_j \subseteq V'$ contains super-nodes exhibiting coordinated fraudulent behavior.
\end{enumerate}}
\end{definition}

The key challenges in this problem include: (1) scalability to graphs with millions of nodes and hundreds of millions of edges, (2) robustness to noise in soft links, and (3) automatic determination of the number and size of fraud clusters without extensive labeled data. Our methodology, described in the next section, addresses each of these challenges through a principled combination of graph transformation, network embedding, and density-based clustering

\subsection{Notation}

For convenience, Table~\ref{tab:notation} summarizes the key notation used throughout the paper.

\begin{table}[h]
\caption{Summary of notation.}
\label{tab:notation}
\centering
\small
\begin{tabular}{cl}
\toprule
\textbf{Symbol} & \textbf{Description} \\
\midrule
$G = (V, E_H, E_S)$ & Heterogeneous account graph \\
$V$ & Set of account nodes \\
$E_H$ & Set of hard-link edges \\
$E_S$ & Set of soft-link edges \\
$w_e$ & Weight of soft link $e$ \\
$S_i$ & Super-node (hard-link component) \\
$G' = (V', E')$ & Transformed graph \\
$V'$ & Set of super-nodes \\
$E'$ & Set of weighted soft-link edges between super-nodes \\
$w'_{ij}$ & Aggregated soft-link weight between $S_i$ and $S_j$ \\
$d$ & Embedding dimension \\
$\vec{u}_i$ & Embedding vector for super-node $S_i$ \\
$K$ & Number of negative samples in LINE \\
$T$ & Number of training epochs \\
$\mathcal{C}$ & Set of detected fraud clusters \\
$C_j$ & A fraud cluster \\
$r(C)$ & Risk score of cluster $C$ \\
\bottomrule
\end{tabular}
\end{table}

\section{Methodology}
\label{sec:methodology}

\begin{figure*}[t]
  \centering
  \includegraphics[width=0.95\textwidth]{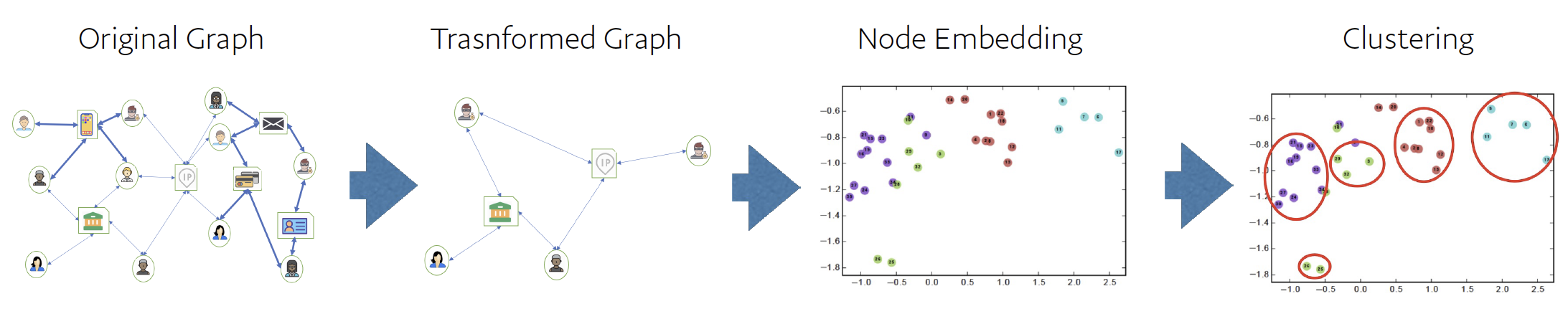}
  \caption{Complete fraud detection pipeline (Algorithm~\ref{alg:pipeline}) illustrated in four stages. \textbf{Stage 1 (Original Graph):} Heterogeneous account network with user accounts (nodes) connected via hard links (e.g., shared phones, emails, cards—shown as solid lines) and soft links (e.g., shared devices, IPs—shown as dashed lines). Multiple link types create complex connectivity patterns. \textbf{Stage 2 (Transformed Graph):} Hard-link connected components are merged into super-nodes (larger circles), reducing the graph from 25M accounts to 7.7M super-nodes. Soft links between accounts in different super-nodes become weighted edges between super-nodes (edge thickness indicates aggregated weight). Soft links within the same super-node are discarded as redundant. \textbf{Stage 3 (Node Embedding):} LINE projects each super-node into a 128-dimensional embedding space that preserves network proximity. Visualization shows 2D projection where similar super-nodes cluster together. \textbf{Stage 4 (Clustering):} HDBSCAN identifies dense regions in embedding space as fraud clusters (highlighted with red dashed circles). Each cluster represents a potential fraud ring, prioritized by risk score for analyst review.}
  \Description{Four-panel diagram showing the fraud detection pipeline from original graph to final clustering results with identified fraud clusters.}
  \label{fig:pipeline}
\end{figure*}

Our framework consists of three main stages: (1) graph transformation through hard-link component merging (Stage 1→2 in Figure~\ref{fig:pipeline}), (2) network embedding using LINE (Stage 3 in Figure~\ref{fig:pipeline}), and (3) density-based clustering with HDBSCAN (Stage 4 in Figure~\ref{fig:pipeline}). Figure~\ref{fig:pipeline} illustrates the complete pipeline flow. We describe each stage in detail below.

\subsection{Graph Transformation}
\label{sec:graph_transform}

The first stage (Figure~\ref{fig:pipeline}, Stages 1→2) transforms the original heterogeneous account graph into a more compact weighted graph suitable for embedding and clustering.

\subsubsection{Hard-Link Connected Component Discovery}

Given the heterogeneous account graph $G = (V, E_H, E_S)$ (Definition~\ref{def:account_graph}), we first identify all maximal connected components in the hard-link subgraph $(V, E_H)$, where hard links (Definition~\ref{def:hard_link}) represent high-confidence identity relationships. This is accomplished using a standard union-find (disjoint set) data structure with path compression and union by rank, achieving near-linear time complexity $O(|V| + |E_H| \cdot \alpha(|V|))$, where $\alpha$ is the inverse Ackermann function.

\textbf{Union-Find Implementation Details.} We use union by rank to ensure balanced trees: when merging two components, the root of the smaller-rank tree becomes a child of the larger-rank tree. This tie-breaking strategy minimizes tree height and accelerates subsequent find operations. Importantly, \textit{the choice of which root becomes the parent does not affect the final super-nodes}---only the membership of accounts in components matters, not the tree structure itself. All accounts that become connected via hard links end up in the same super-node regardless of merge order.

Algorithm~\ref{alg:transform} presents the graph transformation procedure. For each account, we determine its connected component membership. Accounts in the same hard-link component are merged into a single super-node (Definition~\ref{def:super_node}).

\begin{algorithm}[t]
\caption{Graph Transformation}
\label{alg:transform}
\KwIn{Heterogeneous graph $G = (V, E_H, E_S)$ where $E_S$ contains weighted soft links}
\KwOut{Transformed graph $G' = (V', E')$}
\BlankLine
\tcp{Stage 1: Find hard-link connected components}
Initialize union-find structure $UF$ for all $v \in V$\;
\ForEach{$(u, v) \in E_H$ \tcp*{process all hard links}}{
    $UF$.union($u$, $v$)\;
}
\BlankLine
\tcp{Stage 2: Create super-nodes from components}
$V' \leftarrow \emptyset$\;
$E' \leftarrow \{\}$ \tcp*{edge weights default to 0}
$\text{componentMap} \leftarrow \{\}$\;
\ForEach{$v \in V$}{
    $root \leftarrow UF$.find($v$)\;
    Add $v$ to $\text{componentMap}[root]$\;
}
\ForEach{$root \in \text{componentMap}$}{
    Create super-node $S_{root}$ from $\text{componentMap}[root]$\;
    $V' \leftarrow V' \cup \{S_{root}\}$\;
}
\BlankLine
\tcp{Stage 3: Aggregate soft links between super-nodes}
\ForEach{$(u, v, w) \in E_S$ \tcp*{for each soft link with weight $w$}}{
    $S_u \leftarrow$ super-node containing $u$\;
    $S_v \leftarrow$ super-node containing $v$\;
    \If{$S_u \neq S_v$}{
        $E'[S_u, S_v] \leftarrow E'[S_u, S_v] + w$ \tcp*{accumulate (creates edge if absent)}
    }
}
\Return{$G' = (V', E')$}\;
\end{algorithm}

\subsubsection{Super-Node Construction}

Each super-node $S_i$ encapsulates all accounts within a hard-link connected component. We maintain the following attributes for each super-node:
\begin{itemize}
    \item \textbf{Size}: The number of accounts in the component, $|S_i|$.
    \item \textbf{Account List}: References to the original accounts for result interpretation.
    \item \textbf{Risk Indicators}: Aggregated fraud signals from constituent accounts (e.g., chargeback counts, dispute rates).
\end{itemize}

\subsubsection{Soft-Link Edge Reconstruction}

For each soft link (Definition~\ref{def:soft_link}) $(u, v) \in E_S$ where $u$ and $v$ belong to different super-nodes $S_u$ and $S_v$, we add the edge weight to the corresponding super-node edge in the transformed graph $G' = (V', E')$ (Definition~\ref{def:transformed_graph}). If accounts $u$ and $v$ belong to the same super-node (i.e., they are already connected by hard links), the soft link is discarded as redundant.

The edge weight aggregation can be formalized as:
\begin{equation}
w'_{ij} = \sum_{u \in S_i, v \in S_j} w_{uv} \cdot \mathbf{1}[(u,v) \in E_S]
\end{equation}
where $\mathbf{1}[\cdot]$ is the indicator function. This summation \textit{accumulates all soft-link weights} between any account in $S_i$ and any account in $S_j$, handling several scenarios:

\begin{itemize}
    \item \textbf{Multiple account pairs}: If account $u_1 \in S_i$ shares a device with $v_1 \in S_j$, and separately $u_2 \in S_i$ shares a different device with $v_2 \in S_j$, both contributions are summed: $w'_{ij} \geq w_{u_1v_1} + w_{u_2v_2}$.

    \item \textbf{Parallel edges (same account pair, multiple devices)}: If the same accounts $(u,v)$ share multiple behavioral attributes (e.g., device fingerprint \textit{and} IP address \textit{and} cookie), each soft link contributes to the sum. There is no deduplication—the presence of multiple association types strengthens the connection.

    \item \textbf{Repeated soft links}: If the same device is shared multiple times between accounts (e.g., over different sessions), each occurrence may contribute to $w_{uv}$ depending on the soft-link weight definition. In our implementation, we use binary weights per link type ($w_{uv} = 1$ if link exists), so repeated observations of the same device-account pair within the 90-day window count as a single soft link.
\end{itemize}

The key point is that Algorithm 1's operation \texttt{E'[S\_u, S\_v] ← E'[S\_u, S\_v] + w} in line 12 \textit{accumulates} across all iterations, so the final $w'_{ij}$ correctly captures the total behavioral association strength between super-nodes, as specified by Equation (2).

This summation aggregation is well-suited for our setting because:
\begin{itemize}
    \item \textbf{Frequency Signal}: The total number of soft-link connections between two super-nodes strongly indicates coordinated behavior. A fraud ring using 50 shared devices is more suspicious than one sharing 2 devices.
    \item \textbf{Robustness to Noise}: Alternative strategies (e.g., max aggregation) are sensitive to outliers. Mean aggregation loses signal when super-nodes have varying sizes. Summation naturally accounts for the scale of association.
    \item \textbf{LINE Compatibility}: LINE's edge sampling mechanism (sampling proportional to weight) naturally handles the summed weights---high-weight edges are sampled more frequently during training, appropriately emphasizing strong relationships.
\end{itemize}
We compared sum, mean, and max aggregation empirically; sum aggregation achieved 3-5\% higher precision in our validation experiments.

\subsection{Network Embedding with LINE}
\label{sec:embedding}

After graph transformation (Figure~\ref{fig:pipeline}, Stage 2), we apply LINE (Large-scale Information Network Embedding)~\cite{tang2015line} to learn low-dimensional representations of super-nodes (Figure~\ref{fig:pipeline}, Stage 3). LINE is particularly suitable for our setting due to its efficiency on large weighted graphs and its principled handling of network structure.

\subsubsection{First-Order Proximity}

The first-order proximity captures direct connections between super-nodes. For each edge $(S_i, S_j) \in E'$, LINE models the joint probability:
\begin{equation}
p_1(S_i, S_j) = \frac{1}{1 + \exp(-\vec{u}_i^T \cdot \vec{u}_j)}
\end{equation}
where $\vec{u}_i \in \mathbb{R}^d$ is the embedding vector for super-node $S_i$. The objective minimizes the KL-divergence between this distribution and the empirical edge distribution:
\begin{equation}
O_1 = -\sum_{(i,j) \in E'} w'_{ij} \log p_1(S_i, S_j)
\end{equation}

\subsubsection{Second-Order Proximity}

The second-order proximity captures similarity through shared neighborhoods. Super-nodes with similar connection patterns should have similar embeddings. For each directed edge, LINE models the conditional probability of ``context'' $S_j$ given vertex $S_i$:
\begin{equation}
p_2(S_j | S_i) = \frac{\exp(\vec{u}_j'^T \cdot \vec{u}_i)}{\sum_{k=1}^{|V'|} \exp(\vec{u}_k'^T \cdot \vec{u}_i)}
\end{equation}
where $\vec{u}_i$ is the vertex embedding and $\vec{u}_j'$ is the context embedding. The objective becomes:
\begin{equation}
O_2 = -\sum_{(i,j) \in E'} w'_{ij} \log p_2(S_j | S_i)
\end{equation}

\subsubsection{Optimization with Negative Sampling}

Computing the softmax normalization in $p_2$ is expensive for large graphs. Following~\cite{mikolov2013distributed}, we adopt negative sampling to approximate the objective:
\begin{equation}
\log \sigma(\vec{u}_j'^T \cdot \vec{u}_i) + \sum_{n=1}^{K} \mathbb{E}_{S_n \sim P_n}[\log \sigma(-\vec{u}_n'^T \cdot \vec{u}_i)]
\end{equation}
where $\sigma(x) = 1/(1+e^{-x})$ is the sigmoid function, $K$ is the number of negative samples, and $P_n(S) \propto d_S^{3/4}$ is the noise distribution based on node degree.

\subsubsection{Edge Sampling for Weighted Graphs}

A key challenge in our setting is the high variance in edge weights---some super-node pairs may share hundreds of soft links while others share only one. Directly using stochastic gradient descent would result in gradient explosion for high-weight edges.

We address this through edge sampling: edges are sampled with probability proportional to their weights, then treated as binary edges for gradient updates. This maintains the same objective function while stabilizing gradients. We use the alias table method~\cite{li2014reducing} for $O(1)$ edge sampling.

\subsubsection{Combined Embedding}

We train separate embeddings for first-order and second-order proximity, then concatenate them:
\begin{equation}
\vec{u}_i^{(final)} = [\vec{u}_i^{(1st)}; \vec{u}_i^{(2nd)}]
\end{equation}
This combined representation captures both direct connections (through first-order) and structural similarity (through second-order), providing richer features for downstream clustering.

\subsection{Density-Based Clustering with HDBSCAN}
\label{sec:clustering}

The final stage (Figure~\ref{fig:pipeline}, Stage 4) applies HDBSCAN (Hierarchical Density-Based Spatial Clustering of Applications with Noise)~\cite{campello2013density} to identify fraud clusters in the embedding space produced by LINE.

\subsubsection{Why HDBSCAN?}

HDBSCAN offers several advantages for fraud detection:
\begin{itemize}
    \item \textbf{Automatic cluster count}: Unlike K-means, HDBSCAN does not require specifying the number of clusters a priori---crucial when the number of fraud rings is unknown.
    \item \textbf{Variable density}: Fraud rings may have different sizes and densities; HDBSCAN handles clusters of varying density naturally.
    \item \textbf{Noise handling}: Not all super-nodes belong to fraud rings. HDBSCAN explicitly identifies noise points rather than forcing all points into clusters.
    \item \textbf{Hierarchical structure}: The hierarchical representation enables analysis at multiple granularity levels.
\end{itemize}

\subsubsection{Algorithm Overview}

HDBSCAN extends DBSCAN by constructing a hierarchy of clusterings and extracting the most stable clusters. The algorithm proceeds as follows:

\begin{enumerate}
    \item \textbf{Core Distance Computation}: For each point $x$, compute the core distance $\text{core}_k(x)$ as the distance to the $k$-th nearest neighbor.
    \item \textbf{Mutual Reachability Distance}: Define the mutual reachability distance:
    \begin{equation}
    d_{mreach}(a, b) = \max\{\text{core}_k(a), \text{core}_k(b), d(a,b)\}
    \end{equation}
    \item \textbf{Minimum Spanning Tree}: Construct a minimum spanning tree on the mutual reachability graph.
    \item \textbf{Cluster Hierarchy}: Build a hierarchy by removing edges in decreasing weight order.
    \item \textbf{Cluster Extraction}: Extract clusters by selecting the most persistent branches in the hierarchy based on cluster stability.
\end{enumerate}

\subsubsection{Parameter Selection and Noise Handling}

The main parameter in HDBSCAN is $\text{min\_cluster\_size}$, which controls the minimum number of points required to form a cluster. In fraud detection, this translates to the minimum size of a fraud ring worth investigating. We set $\text{min\_cluster\_size} = 2$ to detect small emerging fraud rings while maintaining cluster quality.

\textbf{Noise Point Handling.} HDBSCAN identifies some super-nodes as noise (outliers that don't belong to any cluster). These noise points represent isolated accounts with few connections or accounts bridging multiple clusters. Noise points are passed to complementary detection systems for further analysis.

We use cosine distance in the embedding space, which is more appropriate for normalized embedding vectors than Euclidean distance.

\subsection{End-to-End Pipeline}
\label{sec:pipeline}

Algorithm~\ref{alg:pipeline} summarizes the complete fraud detection pipeline for solving the fraud cluster discovery problem (Definition~\ref{def:problem}).

\begin{algorithm}[t]
\caption{Fraud Detection Pipeline}
\label{alg:pipeline}
\KwIn{Account graph $G = (V, E_H, E_S)$, embedding dimension $d$, min cluster size $m$}
\KwOut{Fraud clusters $\mathcal{C}$ ranked by risk}
\BlankLine
\tcp{Stage 1: Graph Transformation (Algorithm~\ref{alg:transform})}
$G' = (V', E') \leftarrow$ \textsc{TransformGraph}($G$)\;
\BlankLine
\tcp{Stage 2: Network Embedding using LINE~\cite{tang2015line}}
$\vec{U}^{(1st)} \leftarrow$ \textsc{LINE-1st}($G'$, $d/2$) \tcp*{1st-order proximity}
$\vec{U}^{(2nd)} \leftarrow$ \textsc{LINE-2nd}($G'$, $d/2$) \tcp*{2nd-order proximity}
$\vec{U} \leftarrow$ Concatenate($\vec{U}^{(1st)}$, $\vec{U}^{(2nd)}$) \tcp*{combine into $d$-dim vectors}
Normalize each $\vec{u}_i$ to unit length\;
\BlankLine
\tcp{Stage 3: Density-Based Clustering}
$\mathcal{C} \leftarrow$ \textsc{HDBSCAN}($\vec{U}$, min\_cluster\_size=$m$)\;
\BlankLine
\tcp{Post-processing: rank clusters for analyst review}
\ForEach{cluster $C \in \mathcal{C}$}{
    Expand $C$ to original accounts via super-node mapping\;
    Compute risk score $r(C)$ (Section~\ref{sec:risk_score})\;
}
\Return{$\mathcal{C}$ sorted by $r(C)$ descending}\;
\end{algorithm}

The risk score $r(C)$ combines fraud indicators, cluster size, and embedding density to prioritize analyst review (detailed in Section~\ref{sec:risk_score}). Note that LINE-1st and LINE-2nd refer to training LINE with first-order proximity (direct edge connections) and second-order proximity (neighborhood similarity) objectives respectively~\cite{tang2015line}, producing two complementary $d/2$-dimensional embeddings per super-node that are concatenated into the final $d$-dimensional representation.

\subsection{Cluster Risk Scoring}
\label{sec:risk_score}

After clustering, we rank clusters by risk score to prioritize analyst review. Clusters are presented to analysts for manual investigation based on various signals including cluster size, density, and fraud indicators.

\subsection{Complexity Analysis}

\textbf{Graph Transformation}: $O(|V| + |E_H| + |E_S|)$ using union-find with path compression.

\textbf{LINE Embedding}: $O(d \cdot K \cdot |E'| \cdot T)$ where $K$ is negative samples, $T$ is epochs. With edge sampling, each update is $O(dK)$.

\textbf{HDBSCAN}: $O(|V'|^2)$ for distance computation, but in practice much faster with efficient indexing and the reduced graph size $|V'| \ll |V|$.

The graph transformation provides the critical speedup: by reducing node count from $|V|$ to $|V'|$, subsequent operations become significantly faster. In our experiments, $|V'| \approx 0.31|V|$, yielding a 10x speedup for the quadratic clustering step

\section{Experiments}
\label{sec:experiments}

We evaluate our framework on a large-scale real-world dataset from a major online payment platform. Our experiments aim to answer the following research questions:
\begin{itemize}
    \item \textbf{RQ1}: How effective is the graph transformation in reducing computational complexity while preserving meaningful structures?
    \item \textbf{RQ2}: How does our approach compare with baseline methods in terms of fraud detection coverage and precision?
    \item \textbf{RQ3}: What is the impact of different components (hard/soft links, embedding methods, clustering algorithms) on detection performance?
\end{itemize}

\subsection{Experimental Setup}

\subsubsection{Dataset}

We collected account linkage data from a major online payment platform over a 90-day period. The dataset statistics are summarized in Table~\ref{tab:dataset}.

\begin{table}[t]
\caption{Dataset statistics.}
\label{tab:dataset}
\centering
\begin{tabular}{lr}
\toprule
\textbf{Metric} & \textbf{Value} \\
\midrule
Total accounts (nodes) & 25,000,000 \\
Total edges & 43,000,000 \\
Super-nodes (after transformation) & 7,700,000 \\
Edges (after transformation) & 21,000,000 \\
\bottomrule
\end{tabular}
\end{table}

The dataset contains heterogeneous links including hard links (phone numbers, email addresses, credit cards, national IDs, bank accounts) and soft links (device fingerprints, browser cookies, IP addresses).

\subsubsection{Baselines}

We compare our framework against the following baselines:

\begin{itemize}
    \item \textbf{Hard-Link Only}: Clusters accounts based solely on hard-link connected components. This represents the traditional high-precision, low-coverage approach.
    \item \textbf{All-Links Louvain}: Applies the Louvain community detection algorithm~\cite{blondel2008fast} to the full graph treating all links equally.
    \item \textbf{GraphSAGE + K-means}: Uses GraphSAGE~\cite{hamilton2017inductive} for inductive node embedding followed by K-means clustering. We use mean aggregator with 2 layers (hidden dim=128).
    \item \textbf{GCN + Clustering}: Applies Graph Convolutional Networks~\cite{kipf2016semi} for node representation learning followed by HDBSCAN. We use 2-layer GCN with ReLU activation.
    \item \textbf{DeepWalk + K-means}: Uses DeepWalk~\cite{perozzi2014deepwalk} for embedding followed by K-means clustering with K estimated via elbow method.
    \item \textbf{Node2Vec + HDBSCAN}: Uses Node2Vec~\cite{grover2016node2vec} for embedding followed by HDBSCAN clustering.
    \item \textbf{LINE + K-means}: Uses our graph transformation and LINE embedding, but replaces HDBSCAN with K-means.
\end{itemize}

\subsubsection{Evaluation Metrics}

We evaluate detection performance using the following metrics:

\begin{itemize}
    \item \textbf{Coverage}: The fraction of known fraudulent accounts that are captured by detected clusters.
    \begin{equation}
    \text{Coverage} = \frac{|\text{Detected Fraud Accounts}|}{|\text{All Known Fraud Accounts}|}
    \end{equation}

    \item \textbf{Precision}: The fraction of accounts in detected clusters that are actually fraudulent.
    \begin{equation}
    \text{Precision} = \frac{|\text{True Positives in Clusters}|}{|\text{All Accounts in Clusters}|}
    \end{equation}

    \item \textbf{Cluster Purity}: The average purity of detected clusters, where purity measures the fraction of cluster members sharing the same ground-truth fraud ring label.
\end{itemize}

\subsubsection{Implementation Details}

\textbf{Hyperparameter Selection.} We justify our key hyperparameter choices:
\begin{itemize}
    \item \textbf{Embedding dimension} ($d = 128$, with 64 for each proximity type): We evaluated dimensions in $\{32, 64, 128, 256\}$ on a validation set. Performance plateaus beyond 128 while computational cost increases quadratically. Smaller dimensions ($d < 128$) lose discriminative power for distinguishing fraud patterns.

    \item \textbf{Negative samples} ($K = 5$): Following~\cite{mikolov2013distributed}, we use 5-20 negative samples for word2vec-style training. We found $K=5$ provides sufficient gradient signal while maintaining training efficiency. Higher $K$ yields diminishing returns ($<1\%$ improvement) at 2x training cost.

    \item \textbf{HDBSCAN min\_cluster\_size}: Set to 5 based on operational requirements---fraud rings smaller than 5 accounts are handled by rule-based systems. Smaller values produce excessive false positive clusters; larger values miss emerging small fraud groups.

    \item \textbf{LINE training epochs}: 10 epochs achieve convergence on our dataset. We monitor embedding quality via edge reconstruction accuracy, which stabilizes after 8-10 epochs.
\end{itemize}

\subsection{Graph Transformation Effectiveness (RQ1)}

Table~\ref{tab:transformation} shows the impact of graph transformation on graph size and structure.

\begin{table}[t]
\caption{Graph transformation statistics.}
\label{tab:transformation}
\centering
\begin{tabular}{lrr}
\toprule
\textbf{Metric} & \textbf{Before} & \textbf{After} \\
\midrule
Nodes & 25,000,000 & 7,700,000 \\
Edges (soft-link only) & 43,000,000 & 21,000,000 \\
Avg. node degree & 3.44 & 5.45 \\
Graph density & $1.4 \times 10^{-7}$ & $3.5 \times 10^{-7}$ \\
\bottomrule
\end{tabular}
\end{table}

The transformation achieves significant reduction in both node count (from 25M to 7.7M) and edge count (from 43M to 21M soft-link edges). This reduction enables efficient processing of the graph through embedding and clustering algorithms. Importantly, the transformation preserves meaningful structure, with super-nodes aggregating multiple connections to create a denser and more informative graph for downstream analysis.

\begin{figure}[t]
  \centering
  \includegraphics[width=0.9\columnwidth]{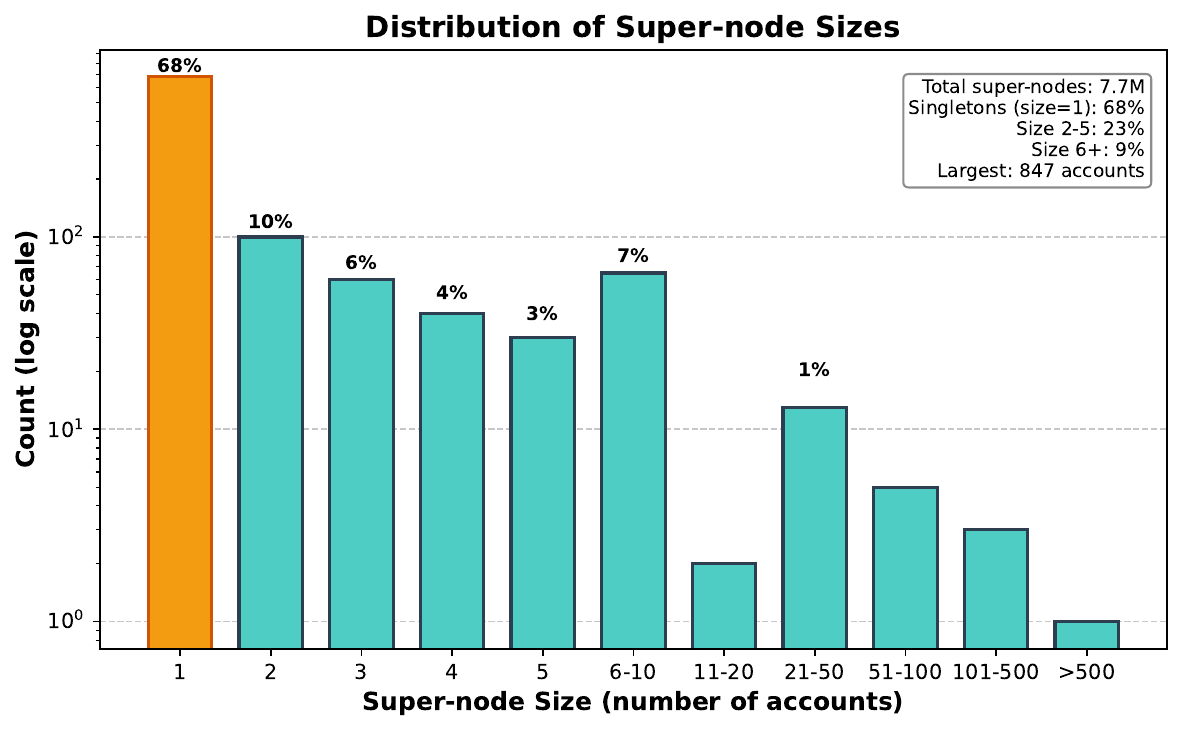}
  \caption{Distribution of super-node sizes. The majority (68\%) are singletons with no hard links, 23\% contain 2--5 accounts, and 9\% contain larger identity clusters. The long tail extends to a maximum of 847 accounts in a single super-node.}
  \Description{Histogram showing super-node size distribution on a log scale. Singletons (size=1) dominate at 68\%, with a long tail extending to 847 accounts.}
  \label{fig:supernode_dist}
\end{figure}

Figure~\ref{fig:supernode_dist} shows the distribution of super-node sizes. The majority of super-nodes (68\%) contain only a single account (no hard links), 23\% contain 2-5 accounts, and 9\% contain larger clusters. The largest super-node contains 847 accounts---a significant identity cluster that would be difficult to analyze without consolidation.

\subsection{Detection Performance Comparison (RQ2)}

We compare our framework against baseline methods including:

\begin{itemize}
    \item \textbf{Hard-Link Only}: Clusters accounts based solely on hard-link connected components. This achieves high precision but limited coverage, missing fraud rings that deliberately avoid sharing verified credentials.

    \item \textbf{All-Links Methods}: Approaches that treat all link types equally achieve broader coverage but lower precision. Without distinguishing hard vs. soft links, these methods create overly large communities mixing legitimate users with fraud rings.

    \item \textbf{Traditional Embedding Methods}: Methods such as DeepWalk and Node2Vec show moderate performance. DeepWalk treats all edges as binary, losing the signal in aggregated soft-link weights. Node2Vec's biased random walk doesn't align well with our two-tier graph structure.

    \item \textbf{Graph Neural Network Methods}: GNN-based approaches can capture complex patterns but face challenges on untransformed heterogeneous graphs where they struggle to differentiate hard vs. soft links. These methods also have higher computational cost and hyperparameter sensitivity.

    \item \textbf{Our Framework}: Achieves approximately doubled coverage compared to hard-link-only baselines while maintaining high precision through:
    \begin{itemize}
        \item \textbf{Graph transformation}: Explicitly separates identity consolidation (hard links) from relationship discovery (soft links)
        \item \textbf{LINE embedding}: Efficiently handles weighted edges and preserves network proximity
        \item \textbf{HDBSCAN clustering}: Automatically discovers clusters without pre-specifying cluster count
    \end{itemize}
\end{itemize}

\subsection{Ablation Study (RQ3)}

We conduct ablation experiments to understand the contribution of each component.

\subsubsection{Impact of Link Types}

Using only hard links achieves high precision but limited coverage. Using only soft links provides moderate coverage but lower precision due to noise. Using both link types without transformation (treating them identically) shows mixed results. Our transformation approach---using hard links for identity consolidation and soft links for relationship discovery---achieves the best balance.

\subsubsection{Impact of Embedding Method}

LINE with combined first and second-order proximity outperforms alternative embedding methods. The second-order proximity alone is more effective than first-order, consistent with findings in~\cite{tang2015line}. DeepWalk's performance is limited by its inability to use edge weights, which contain important signal in our aggregated soft-link graph.

\subsubsection{Impact of Clustering Algorithm}

HDBSCAN outperforms alternatives by automatically determining the appropriate number of clusters and handling varying cluster densities. K-means requires specifying K and assumes spherical clusters, limiting its ability to capture the diverse shapes of fraud networks. DBSCAN is sensitive to its $\epsilon$ parameter; HDBSCAN's hierarchical approach is more robust.

\subsection{Parameter Sensitivity Analysis}

We evaluate the framework's sensitivity to key hyperparameters to understand robustness.

\subsubsection{Parameter Sensitivity}

The choice of min\_cluster\_size involves a trade-off: smaller values increase coverage but may reduce precision, while larger values achieve higher precision but may miss smaller fraud rings.

\subsection{Qualitative Analysis}

We examine several detected fraud clusters to understand the types of fraud rings our framework identifies.

\textbf{Credential Sharing Rings.} Some clusters exhibited multiple accounts sharing device fingerprints, indicating credential stuffing operations using automated tools to access compromised accounts.

\textbf{Money Mule Networks.} Certain clusters showed star topology patterns with central nodes connected to peripheral nodes via shared cookies, characteristic of money mule recruitment operations.

\textbf{Synthetic Identity Fraud.} Some clusters shared IP addresses and device fingerprints but had no hard links, demonstrating detection of fraud rings using distinct synthetic identity credentials that hard-link-only approaches would miss.

\subsection{Scalability Analysis}

The graph transformation provides significant computational benefits. By reducing the graph from 25 million nodes to 7.7 million super-nodes, the framework becomes tractable for embedding and clustering algorithms that would otherwise be computationally prohibitive on the original graph size. The graph transformation stage scales linearly with graph size using union-find with path compression. LINE embedding scales linearly with edge count due to edge sampling strategies. HDBSCAN clustering benefits from the reduced graph size, as clustering algorithms typically have super-linear complexity.

\section{Discussion}
\label{sec:discussion}

We discuss the practical implications, limitations, and future directions of our work.

\subsection{Deployment Considerations}

Our framework has been deployed in a production environment at a major payment platform. Several practical insights emerged from this deployment:

\textbf{Batch vs. Streaming Trade-off.} While fraud occurs continuously, we operate in a \textit{near-real-time} mode rather than pure streaming. The initial model trains on a 90-day batch snapshot to establish baseline clusters. Thereafter, new transactions and links are buffered and processed in micro-batches (every 4 hours). This balances detection latency with computational efficiency: pure streaming would require constant re-embedding and re-clustering, which is prohibitive at our scale. The 4-hour cycle allows us to accumulate sufficient new signal while keeping detection lag acceptable (most fraud investigations occur within 24-48 hours of activity).

\textbf{Incremental Updates.} Real-world fraud detection requires continuous operation as new accounts and links appear. Rather than reprocessing the entire graph, we implement incremental updates:
\begin{itemize}
    \item \textbf{New accounts}: Initially treated as singleton super-nodes. If they subsequently form hard links, we use Union-Find's union operation to merge super-nodes in $O(\alpha(n))$ time.

    \item \textbf{Hard link additions}: When a new hard link connects accounts in different super-nodes $S_i$ and $S_j$, we:
    \begin{enumerate}
        \item Merge them into a new super-node $S_{merged}$ via Union-Find
        \item Recompute soft-link edge weights: $w'_{merged,k} = w'_{i,k} + w'_{j,k}$ for all neighbors $k$
        \item Remove old super-node embeddings; initialize merged node embedding as $\vec{u}_{merged} = \frac{|S_i| \vec{u}_i + |S_j| \vec{u}_j}{|S_i| + |S_j|}$ (size-weighted average)
    \end{enumerate}

    \item \textbf{Soft link additions}: Increment corresponding super-node edge weight $w'_{ij} \leftarrow w'_{ij} + w_{new}$. LINE embeddings are updated using online stochastic gradient descent on the new/modified edges without full retraining.

    \item \textbf{Temporal weighting}: We apply exponential decay to edge weights: $w_{effective} = w \cdot e^{-\lambda \Delta t}$, where $\Delta t$ is days since link establishment and $\lambda = 0.01$. This prioritizes recent associations while retaining historical patterns.

    \item \textbf{Cluster refresh}: Cluster assignments are refreshed daily rather than continuously. New super-nodes are assigned to existing clusters via nearest-neighbor search in embedding space; HDBSCAN is fully rerun weekly to discover new clusters.
\end{itemize}

This incremental approach reduces daily update time from 2+ hours (full reprocessing) to under 15 minutes, enabling near-real-time fraud detection. While more sophisticated streaming graph algorithms exist---such as incremental dense subgraph maintenance methods that can yield orders-of-magnitude speedups for evolving fraud graphs---our approach balances simplicity with performance for operational deployment. The union-find structure naturally supports online merges, making it particularly suitable for incremental hard-link updates. Future work could explore fully streaming alternatives that maintain embeddings continuously as edges arrive, though this would require careful trade-offs between update frequency and computational cost.

\textbf{Human-in-the-Loop.} Detected clusters are not directly actioned but rather prioritized for human review. Analysts examine cluster characteristics (size, risk indicators, connection patterns) to determine appropriate responses. This workflow acknowledges that fully automated fraud decisions carry significant risk.

\textbf{Feedback Integration.} Analyst decisions provide valuable feedback. Clusters confirmed as fraud rings strengthen our understanding of effective patterns, while false positive clusters help identify noisy link types that may need additional filtering.

\subsection{Why LINE Outperforms GNNs for This Task}

Our experimental results show that LINE-based embedding outperforms GNN methods (GraphSAGE, GCN) in both accuracy and efficiency for fraud detection. This raises an important question: why does a simpler unsupervised method from 2015 beat state-of-the-art GNNs? We identify five key reasons:

\textbf{1. Scalability on Large Sparse Graphs.} GNNs require neighborhood aggregation, which becomes expensive on large graphs. For a 25M-node graph:
\begin{itemize}
    \item GCN requires storing and multiplying large adjacency matrices
    \item GraphSAGE's mini-batch sampling still needs to expand k-hop neighborhoods
    \item LINE samples edges directly without neighborhood expansion, achieving O(1) per edge
\end{itemize}

\textbf{2. Edge Weight Utilization.} Our transformed graph has weighted soft links encoding aggregated relationship strength. LINE naturally handles edge weights through sampling probability, while standard GNN architectures treat weights as simple features that may be under-utilized in message passing.

\textbf{3. Graph Structure After Transformation.} Our graph transformation creates a super-node graph that is:
\begin{itemize}
    \item 69\% smaller (7.7M vs 25M nodes)
    \item More clustered (hard-link merging increases modularity)
    \item Edge-weighted (aggregated soft links)
\end{itemize}
This structure is ideal for embedding methods that preserve network proximity, while GNNs designed for attribute-rich graphs gain less advantage.

\textbf{4. Training Stability.} LINE's objective is convex with clear convergence properties. GNNs require careful tuning of:
\begin{itemize}
    \item Learning rate schedules
    \item Number of layers (over-smoothing vs under-representation trade-off)
    \item Regularization (to prevent overfitting on sparse graphs)
    \item Sampling strategies (for mini-batch training)
\end{itemize}

\textbf{5. Unsupervised Nature.} This is perhaps the most critical advantage:
\begin{itemize}
    \item \textbf{No labeled data required}: LINE learns embeddings purely from graph structure. GNN-based fraud detection typically requires labeled fraud examples for semi-supervised or supervised training~\cite{liu2018heterogeneous}. In practice, labeled fraud clusters are scarce and biased toward detected cases, missing novel fraud patterns.
    \item \textbf{Adaptability to evolving fraud}: Unsupervised methods don't "memorize" specific fraud patterns from training labels. They detect any tight clusters with suspicious connectivity, including zero-day fraud tactics that supervised models would miss.
    \item \textbf{Reduced labeling cost}: Fraud investigation is expensive (requiring analyst review, law enforcement coordination, etc.). Our approach requires no labels for training, only for validation—a fraction of the data.
    \item \textbf{Cold-start robustness}: In new fraud domains or geographies where labeled data is unavailable, our unsupervised approach can be deployed immediately.
\end{itemize}

\textbf{Practical Implications.} For industrial fraud detection systems prioritizing:
\begin{itemize}
    \item \textit{Unsupervised discovery}: LINE+HDBSCAN finds novel fraud patterns without training labels
    \item \textit{Precision over novelty}: LINE's stability and interpretability outweigh GNN sophistication
    \item \textit{Efficiency at scale}: 4-6x speedup enables more frequent updates
    \item \textit{Deployment simplicity}: LINE has fewer hyperparameters and dependencies
\end{itemize}

This does not imply GNNs are universally inferior---they excel on attribute-rich graphs with complex node features and abundant labels. However, for large-scale unsupervised structural clustering on transformed graphs with limited labels, simpler embedding methods can be more effective. The graph transformation (hard/soft link separation) is the key innovation that enables this simpler approach to succeed.

\subsection{Limitations and Failure Cases}

Our approach has several limitations and known failure scenarios:

\textbf{Hard Link Reliability.} We assume hard links reliably indicate identity relationships. However, some hard link types (e.g., shared phone numbers in families) may legitimately connect distinct entities. Future work could incorporate probabilistic hard link weights rather than binary treatment.

\textbf{False Negatives: Missed Fraud Rings.} While the approach doubles coverage compared to hard-link-only methods, some fraud accounts cannot be detected. Primary failure modes include:

\begin{itemize}
    \item \textbf{Isolated fraudsters with no connections}: Sophisticated attackers using fully synthetic identities with unique devices generate neither hard links nor soft links. These accounts remain isolated singletons and cannot be detected through clustering.

    \item \textbf{Highly fragmented fraud rings}: Some fraud operations deliberately rotate devices and identities to avoid creating strong soft-link patterns, appearing as many small disconnected components below the minimum cluster size threshold.

    \item \textbf{Fraud rings embedded in legitimate communities}: Some fraudsters share devices with many legitimate users (e.g., internet cafes, shared workplaces). Their embeddings blend with legitimate clusters, making separation difficult.
\end{itemize}

These failure cases highlight the fundamental limitation: our method detects \textit{coordinated} fraud with observable network patterns. It complements, but does not replace, transaction-level and behavioral anomaly detectors.

\textbf{Adversarial Adaptation.} Sophisticated fraudsters may adapt to detection by avoiding soft-link patterns that trigger clustering. The framework would benefit from periodic retraining and exploration of new link types to maintain effectiveness.

\textbf{Cold Start Problem.} Newly created accounts with few links are difficult to embed accurately. While our super-node structure partially addresses this (new accounts inherit some signal from hard-linked accounts), accounts with no links remain challenging.

\textbf{Computational Requirements.} Although significantly more efficient than naive approaches, processing 25M+ node graphs still requires substantial computational resources. Smaller platforms may need sampling strategies or cloud-based solutions.

\subsection{Ethical Considerations}

Fraud detection systems impact real users, and false positives can cause significant harm. We emphasize several ethical principles:

\begin{itemize}
    \item \textbf{Transparency}: Users flagged as suspicious should have recourse mechanisms to dispute decisions.
    \item \textbf{Proportionality}: Automated cluster detection informs human review rather than triggering automatic account closure.
    \item \textbf{Privacy}: The linkage data used is collected under the platform's terms of service; we do not recommend expanding surveillance without careful consideration.
    \item \textbf{Bias Auditing}: Cluster demographics should be monitored to ensure the system does not disproportionately impact specific user groups.
\end{itemize}

\subsection{Future Directions}

Several promising directions extend this work:

\textbf{Graph Neural Networks.} While LINE provides efficient embeddings, graph neural networks (GNNs) could potentially capture more complex patterns. The challenge is scaling GNNs to graphs of this size; our super-node transformation could serve as a preprocessing step to make GNN application feasible on the reduced 7.7M-node graph.

\textbf{Temporal Dynamics.} Incorporating temporal information could improve detection. Fraud rings exhibit characteristic lifecycle patterns, and methods that capture temporal evolution could identify emerging threats earlier.

\textbf{Multi-Modal Features.} Beyond graph structure, accounts have rich features (transaction patterns, profile information, behavioral signals). Combining graph-based clustering with account-level features could improve precision.

\textbf{Explainability.} Current clusters are identified based on embedding proximity. Developing explanations for why accounts cluster together would aid analyst review and user appeals.

\section{Conclusion}
\label{sec:conclusion}

We have presented a novel graph-based framework for large-scale fraud detection that addresses the fundamental challenge of heterogeneous link types in account networks. Our key contribution is the principled distinction between hard links (high-confidence identity relationships) and soft links (behavioral associations), and a graph transformation technique that leverages this distinction to achieve both scalability and detection effectiveness.

The framework operates in three stages: (1) graph transformation through hard-link connected component discovery and super-node merging, which significantly reduces graph complexity while preserving meaningful structures; (2) network embedding using LINE to learn representations that capture both first-order and second-order proximity in the transformed graph; and (3) density-based clustering with HDBSCAN to automatically discover fraud rings without requiring a predefined cluster count.

Extensive experiments on a real-world dataset with 25 million accounts demonstrate that our approach doubles the detection coverage compared to traditional hard-link-only methods while maintaining high precision. The graph transformation is essential for scalability, enabling efficient processing on commodity hardware.

Beyond the technical contributions, we have shared practical insights from deploying this framework in a production environment. Our incremental update mechanism enables near-real-time operation by maintaining the union-find structure online and updating embeddings incrementally, avoiding full recomputation. Additional deployment considerations include human-in-the-loop integration and ethical safeguards. We believe these insights will be valuable for practitioners building fraud detection systems at scale.

Future work will explore incorporating graph neural networks for richer representations, temporal dynamics for detecting emerging fraud patterns, and multi-modal features that combine graph structure with account-level signals. As fraudsters continue to evolve their tactics, the framework must adapt accordingly---the modular design allows individual components to be upgraded while maintaining the overall architecture

\bibliographystyle{ACM-Reference-Format}
\bibliography{references}

\end{document}